# Learning Structural Changes of Gaussian Graphical Models in Controlled Experiments

**Bai Zhang and Yue Wang**
Bradley Department of Electrical and Computer Engineering
Virginia Polytechnic Institute and State University
Arlington, VA 22203

## Abstract

Graphical models are widely used in scientific and engineering research to represent conditional independence structures between random variables. In many controlled experiments, environmental changes or external stimuli can often alter the conditional dependence between the random variables, and potentially produce significant structural changes in the corresponding graphical models. Therefore, it is of great importance to be able to detect such structural changes from data, so as to gain novel insights into where and how the structural changes take place and help the system adapt to the new environment. Here we report an effective learning strategy to extract structural changes in Gaussian graphical model using $\ell_1$-regularization based convex optimization. We discuss the properties of the problem formulation and introduce an efficient implementation by the block coordinate descent algorithm. We demonstrate the principle of the approach on a numerical simulation experiment, and we then apply the algorithm to the modeling of gene regulatory networks under different conditions and obtain promising yet biologically plausible results.

## 1 Introduction

Controlled experiments are very common yet effective tools in scientific research. For example, in the studies of disease or drug effectiveness using case-control experiments, the changes of the conditional dependence between measurement variables are often reflected in the structural changes in the corresponding graphical models that can reveal crucial information about how the systems responds to external stimuli or adapts to changed conditions. The ability to detect and extract the structural changes from data can facilitate the generation of new insights and new hypotheses for further studies.

Consider the example of gene regulatory networks in systems biology. Gene regulatory networks are context-specific and dynamic in nature, that is, under different conditions, different regulatory components and mechanisms are activated and accordingly the topology of the underlying gene regulatory network changes (Zhang *et al.*, 2009). For example, in response to diverse conditions in the yeast, transcription factors alter their interactions and rewire the signaling networks (Luscombe *et al.*, 2004). Such changes in network structures provide great insights into the underlying biology of how the organism responses to outside stimuli. In disease studies, it is important to examine the topological changes in transcriptional networks between disease and normal conditions where a deviation from normal regulatory network topology may reveal the mechanism of pathogenesis, and the genes that undergo the most network topological changes may serve as potential biomarkers or drug targets.

Similar phenomena also appear in other areas. For instance, in web search or collaborative filtering, useful information can be acquired by observing how certain events (*e.g.*, launch of an advertisement campaign) trigger changes in dependence patterns of search keywords or preference for products reflected in the associated structural changes.

The conditional dependence of a set of random variables are often mathematically characterized by graphical models, such as Bayesian networks and Markov networks, and various methods have been proposed to learn graphical model structures from data (Lauritzen, 1996; Jordan, 1998). Although learning the graphical models under two conditions can be separately achieved and the structural and parametric differences can be subsequently compared, such technically convenient framework would completely collapse when the structural and parametric inconsistencies due to limited data samples and noise effects are significant and hinder an accurate detection of true and meaningful structural and parametric changes.

Here we report an effective learning strategy to extract structural changes of Gaussian graphical models in controlled experiments using convex optimization. We discuss the properties of the problem formulation and introduce an efficient block coordinate descent algorithm. We demonstrate the principle of the approach on a numerical simulation experiment, and we then apply the algorithm to the modeling of gene regulatory networks under different conditions and obtain promising yet biologically plausible results.

## 2 A Revisit on Gaussian Graphical Model Structural Learning Using $\ell_1$-regularization

The structures of graphical models in many cases are unknown and need to be learned from data. In this paper, we focus on Gaussian graphical models, in which the nodes (variables) are Gaussian, and their dependence relationships are linear. Assume we have a set of $p$ random variables of interest, $\mathbb{X} = \{X_1, X_2, ..., X_p\}$, and $N$ observations, $\boldsymbol{x}_j = [x_{1j}, x_{2j}, ..., x_{Nj}]^T$, $j = 1, 2, ..., p$. Let $\mathbf{X} = [\boldsymbol{x}_1, \boldsymbol{x}_2, ..., \boldsymbol{x}_p]$ be the data matrix.

Learning the structures of graphical models efficiently is often very challenging. Recently, $\ell_1$-regularization has drawn great interest in statistics and machine learning community (Tibshirani, 1996; Efron *et al.*, 2004; Zou and Hastie, 2005; Zhao and Yu, 2006). Penalty on the $\ell_1$-norm of the regression coefficients has two very useful properties: sparsity and convexity. The $\ell_1$-norm penalty tends to make some coefficients exactly zeros, leading to a parsimonious solution, which naturally performs variable selection or sparse linear model estimation. Further, the convex nature of $\ell_1$-norm penalty makes the problem computationally tractable, which can be solved readily by many existing convex optimization methods.

Several $\ell_1$-regularization approaches have been successfully applied to graphical model structure learning (Lee *et al.*, 2006; Wainwright *et al.*, 2006; Schmidt *et al.*, 2007), especially Gaussian graphical models (Meinshausen and Bühlmann, 2006; Banerjee *et al.*, 2008; Friedman *et al.*, 2008). Meinshausen and Bühlmann (2006) proposed to use lasso to identify the neighborhood of the nodes in Gaussian graphs. The neighborhood selection of a node $X_j$, $j = 1, 2, ..., p$, is solved by applying lasso to learn the prediction model of variable $X_j$, given all remaining variables $\mathbb{X}_{-j}$.

The lasso estimate $\hat{\boldsymbol{\beta}}$ is given by

$$\hat{\boldsymbol{\beta}} = \arg\min_{\boldsymbol{\beta}:\beta_j=0} \|\boldsymbol{x}_j - \mathbf{X}\boldsymbol{\beta}\|_2^2 + \lambda\|\boldsymbol{\beta}\|_1, \qquad (1)$$

where $\lambda > 0$ is the Lagrange multiplier.

If the $k^{\text{th}}$ element of $\hat{\boldsymbol{\beta}}$ is non-zero, then there is an edge between node $j$ and node $k$. This procedure is performed on each of the $p$ random variables, and thereby the structure of the Gaussian graphical model is learned. Meinshausen and Bühlmann (2006) also showed that under certain conditions, the proposed neighborhood selection scheme is consistent for sparse high-dimensional graphs.

## 3 Problem Formulation

Now we consider the problem of learning structural changes of a graphical model between two conditions. This is equivalent to investigating how the conditional dependence and independence of a set of random variables change under these two conditions. Similarly, we have a set of $p$ random variables of interest, $\mathbb{X} = \{X_1, X_2, ..., X_p\}$, and we observed $N_1$ samples under condition 1 and $N_2$ samples under condition 2. Without loss of generality, we assume $N_1 = N_2 = N$, which means we have balanced observations from two conditions. Under the first condition, for variable $X_j$, we have observations $\boldsymbol{x}_j^{(1)} = [x_{1j}^{(1)}, x_{2j}^{(1)}, ..., x_{Nj}^{(1)}]^T$, $j = 1, 2, ..., p$, while under the second condition, we have $\boldsymbol{x}_j^{(2)} = [x_{1j}^{(2)}, x_{2j}^{(2)}, ..., x_{Nj}^{(2)}]^T$, $j = 1, 2, ..., p$. Further, let $\mathbf{X}^{(1)} = [\boldsymbol{x}_1^{(1)}, \boldsymbol{x}_2^{(1)}, ..., \boldsymbol{x}_p^{(1)}]$ be the data matrix under condition 1 and $\mathbf{X}^{(2)} = [\boldsymbol{x}_1^{(2)}, \boldsymbol{x}_2^{(2)}, ..., \boldsymbol{x}_p^{(2)}]$ be the data matrix under condition 2.

Further, denote

$$\boldsymbol{y}_j = \begin{bmatrix} \boldsymbol{x}_j^{(1)} \\ \boldsymbol{x}_j^{(2)} \end{bmatrix}, \qquad (2)$$

$$\mathbf{X} = \begin{bmatrix} \mathbf{X}^{(1)} & \mathbf{0} \\ \mathbf{0} & \mathbf{X}^{(2)} \end{bmatrix}, \qquad (3)$$

and

$$\boldsymbol{\beta} = \begin{bmatrix} \boldsymbol{\beta}^{(1)} \\ \boldsymbol{\beta}^{(2)} \end{bmatrix}$$
$$= [\beta_1^{(1)}, \beta_2^{(1)}, ..., \beta_p^{(1)}, \beta_1^{(2)}, \beta_2^{(2)}, ..., \beta_p^{(2)}]^T. \qquad (4)$$

By location and scale transformations, we can always assume that the variables have mean 0 and unit length,

$$\sum_{i=1}^N x_{ij}^{(1)} = 0, \qquad \sum_{i=1}^N (x_{ij}^{(1)})^2 = 1,$$
$$\sum_{i=1}^N x_{ij}^{(2)} = 0, \qquad \sum_{i=1}^N (x_{ij}^{(2)})^2 = 1, \qquad (5)$$

where $j = 1, 2, ..., p$.

We formulate the problem of learning structural changes between two conditions as a convex optimization problem. We solve the following optimization problem for each node (variable) $X_j$, $j = 1, 2, ..., p$.

$$f(\boldsymbol{\beta}) = \frac{1}{2}\|\boldsymbol{y}_j - \mathbf{X}\boldsymbol{\beta}\|_2^2 + \lambda_1\|\boldsymbol{\beta}\|_1 + \lambda_2\|\boldsymbol{\beta}^{(1)} - \boldsymbol{\beta}^{(2)}\|_1 \quad (6)$$

$$\begin{aligned}
\hat{\boldsymbol{\beta}} &= \arg\min_{\boldsymbol{\beta}} f(\boldsymbol{\beta}) \\
&= \arg\min_{\boldsymbol{\beta}^{(1)}, \boldsymbol{\beta}^{(2)}} \frac{1}{2}\|\boldsymbol{y}_j - \mathbf{X}\boldsymbol{\beta}\|_2^2 + \lambda_1\|\boldsymbol{\beta}\|_1 \\
&\qquad + \lambda_2\|\boldsymbol{\beta}^{(1)} - \boldsymbol{\beta}^{(2)}\|_1 \\
\text{s.t. } & \beta_j^{(1)} = 0, \beta_j^{(2)} = 0
\end{aligned} \quad (7)$$

In (7), we learn the structures of the graphical model under two conditions jointly. The $\ell_2$-loss function and the first $\ell_1$-regularization term, $\lambda_1\|\boldsymbol{\beta}\|_1$, lead to the identification of sparse graph structure. The second $\ell_1$-regularization term, $\lambda_2\|\boldsymbol{\beta}^{(1)} - \boldsymbol{\beta}^{(2)}\|_1$, encourages sparse changes in the model structure and parameters between two conditions, and thereby suppresses the structural and parametric inconsistencies due to noise in the data and limited samples. The objective function (6) is non-differentiable, continuous, and convex.

The optimization problem (7) may appear similar to the fused lasso (Tibshirani *et al.*, 2005), which was applied to protein mass spectroscopy and DNA copy number detection (Friedman *et al.*, 2007). The fused lasso encourages the flatness of the coefficient profile $\beta_j$ as a function of the index $j$. Kolar *et al.* (2009) investigated learning of varying-coefficient varying-structure models from time-course data, in which $\boldsymbol{\beta}_t$ is a function of time $t$, and proposed a two-stage procedure that first identifies jump points and then identifies relevant covariates. The total variation norm (TV-norm) of $\boldsymbol{\beta}_t$ is used to encourage sparse changes along the time course.

Besides targeted at different applications, the objective function (6) has two important technical differences from the above two approaches. First, the penalty term $\lambda_1\|\boldsymbol{\beta}\|_1 + \lambda_2\|\boldsymbol{\beta}^{(1)} - \boldsymbol{\beta}^{(2)}\|_1$ has *block-wise separability*, which means the non-differentiable objective function $f(\boldsymbol{\beta})$ can be written in the form

$$f(\boldsymbol{\beta}) = g(\boldsymbol{\beta}) + \sum_{m=1}^{M} h_m(\boldsymbol{b}_m), \quad (8)$$

where $g(\boldsymbol{\beta})$ is convex and differentiable, $\boldsymbol{b}_m$ is some subset of $\boldsymbol{\beta}$, $h_m(\boldsymbol{b}_m)$ is convex and non-differentiable, and $\boldsymbol{b}_{m_1}$ and $\boldsymbol{b}_{m_2}$, $m_1 \neq m_2$, do not have overlapping members (Tseng, 2001).

We rewrite the objective function (6) as

$$\begin{aligned}
&f(\boldsymbol{\beta}) \\
&= \frac{1}{2}\|\boldsymbol{y}_j - \mathbf{X}\boldsymbol{\beta}\|_2^2 + \lambda_1\sum_{k=1}^{p}(|\beta_k^{(1)}| + |\beta_k^{(2)}|) \\
&\quad + \lambda_2\sum_{k=1}^{p}(|\beta_k^{(1)} - \beta_k^{(2)}|)
\end{aligned}$$

Therefore, the non-differentiable part of $f(\boldsymbol{\beta})$ can be written as the sum of $p$ terms with non-overlapping members, $(\beta_k^{(1)}, \beta_k^{(2)})$, $k = 1, 2, ..., p$. Each $(\beta_k^{(1)}, \beta_k^{(2)})$, $k = 1, 2, ..., p$, is a coordinate block.

We will show in Section 4 that this property is essential for the convergence of the block coordinate descent algorithm to solve problem (7). On the other hand, Friedman *et al.* (2007) has shown that coordinate-wise descent does not work in fused lasso, since the non-differentiable penalty function is not separable.

Additionally, the $k^{\text{th}}$ column of matrix $\mathbf{X}$, $\boldsymbol{x}_k$, and the $(k+p)^{\text{th}}$ column of $\mathbf{X}$, $\boldsymbol{x}_{k+p}$, are orthogonal, *i.e.*, $\boldsymbol{x}_k^T \cdot \boldsymbol{x}_{k+p} = 0$, $k = 1, 2, ..., p$. This simplifies the derivation of closed-form solutions to the sub-problems in each iterations of the block coordinate descent.

We summarize our discussions above as three properties of problem (7).

**Property 1** (Convexity). *The objective function (6) is continuous and convex.*

**Property 2** (Block-wise Separability). *The non-differential part of the objective function (6), $\lambda_1\|\boldsymbol{\beta}\|_1 + \lambda_2\|\boldsymbol{\beta}^{(1)} - \boldsymbol{\beta}^{(2)}\|_1$, is block-wise separable.*

**Property 3** (Orthogonality in the Coordinate Block). $\boldsymbol{x}_k^T \cdot \boldsymbol{x}_{k+p} = 0$, $k = 1, 2, ..., p$.

To represent the result as a graph, the non-zero elements of $\boldsymbol{\beta}^{(1)}$ indicate the neighbors and edges of node $X_j$ under the first condition and the non-zero elements of $\boldsymbol{\beta}^{(2)}$ indicate the neighbors and edges of node $X_j$ under the second condition. The non-zero elements of $\boldsymbol{\beta}^{(1)} - \boldsymbol{\beta}^{(2)}$ provide the changed edges (both structural and parametric difference) of node $X_j$ between two conditions. We repeat this procedure to each node $X_j$, $j = 1, 2, ..., p$, and then we obtain the graph under two conditions. In gene regulatory network modeling, we are particularly interested in where and how the gene regulatory network exhibits different network topology between two conditions. To highlight such changes, we extract the sub-network in which nodes have different connections between two conditions.

## 4 Algorithm

In the realm of computational biology and data mining, vast amount of data and high dimensionality re-quire efficient algorithms. Although the optimization problems with $\ell_1$-regularization can be solved readily by existing convex optimization techniques, a lot of efforts have been made to solve the problems efficiently by exploiting the special structures of the problems. A well-known approach is the least angle regression (LARS), which can be modified to solve lasso problems (Efron *et al.*, 2004). Recently, coordinate-wise descent algorithms have been studied in lasso related problems, such as lasso, garotte and elastic net (Friedman *et al.*, 2007). Friedman *et al.* (2008) showed with experiments that a coordinate descent procedure for lasso, graphical lasso, is 30-4000 times faster than competing methods, making it a computationally attractive method.

### 4.1 Block coordinate descent algorithm

In this paper, we adopt this idea and propose a block coordinate descent algorithm to solve the optimization problem (7) for each node $X_j$, $j = 1, 2, ..., p$. The essence of the block coordinate descent algorithm is "one-block-at-a-time". At iteration $r + 1$, only one coordinate block, $(\beta_k^{(1)}, \beta_k^{(2)})$, is updated, with the remaining $(\beta_l^{(1)}, \beta_l^{(2)})$, $l \neq k$, fixed at their values at iteration $r$. Given

$$\boldsymbol{\beta}^r = [\beta_1^{(1),r}, \beta_2^{(1),r}, ..., \beta_p^{(1),r}, \beta_1^{(2),r}, \beta_2^{(2),r}, ..., \beta_p^{(2),r}]^T, \quad (9)$$

at iteration $r + 1$, the estimation is updated according to the following sub-problem

$$\boldsymbol{\beta}^{r+1} = \arg\min_{\boldsymbol{\beta}} f(\boldsymbol{\beta})$$
$$\text{s.t. } \beta_l^{(1)} = \beta_l^{(1),r},$$
$$\beta_l^{(2)} = \beta_l^{(2),r},$$
$$\text{for } l = 1, 2, ..., p, l \neq k. \quad (10)$$

We use a *cyclic rule* to update parameter estimation iteratively, *i.e.*, update parameter pair $(\beta_k^{(1)}, \beta_k^{(2)})$ at iteration $r + 1$, and $k = ((r + 1) \mod p) + 1$.

### 4.2 Closed-form solution to the sub-problem

Thus the problem is reduced to solving the sub-problem (10). Since $\beta_l^{(1)}$ and $\beta_l^{(2)}$, $l = 1, 2, ..., p, l \neq k$, are fixed during iteration $r+1$, we rewrite the objective function of (10) as

$$\tilde{f}(\boldsymbol{\beta})$$
$$= \frac{1}{2} \| \boldsymbol{y}_j - \sum_{l \neq j,k} \boldsymbol{x}_l \beta_l^{(1),r} - \sum_{l \neq j,k} \boldsymbol{x}_{(p+l)} \beta_l^{(2),r}$$
$$\quad - \boldsymbol{x}_k \beta_k^{(1)} - \boldsymbol{x}_{p+k} \beta_k^{(2)} \|_2^2$$
$$+ \lambda_1 \sum_{l \neq j,k} (|\beta_l^{(1),r}| + |\beta_l^{(2),r}|) + \lambda_2 \sum_{l \neq j,k} (|\beta_l^{(1),r} - \beta_l^{(2),r}|)$$
$$+ \lambda_1 (|\beta_k^{(1)}| + |\beta_k^{(2)}|) + \lambda_2 (|\beta_k^{(1)} - \beta_k^{(2)}|) \quad (11)$$

Let

$$\tilde{\boldsymbol{y}}_j = \boldsymbol{y}_j - \sum_{l \neq j,k} \boldsymbol{x}_l \beta_l^{(1),r} - \sum_{l \neq j,k} \boldsymbol{x}_{(p+l)} \beta_l^{(2),r}. \quad (12)$$

Therefore, updating $(\beta_k^{(1)}, \beta_k^{(2)})$ is equivalent to

$$(\beta_k^{(1),r+1}, \beta_k^{(2),r+1})$$
$$= \arg\min_{\beta_k^{(1)}, \beta_k^{(2)}} \tilde{f}(\boldsymbol{\beta})$$
$$= \arg\min_{\beta_k^{(1)}, \beta_k^{(2)}} \frac{1}{2} \| \tilde{\boldsymbol{y}}_j - \boldsymbol{x}_k \beta_k^{(1)} - \boldsymbol{x}_{p+k} \beta_k^{(2)} \|_2^2$$
$$+ \lambda_1 (|\beta_k^{(1)}| + |\beta_k^{(2)}|) + \lambda_2 (|\beta_k^{(1)} - \beta_k^{(2)}|)$$
$$\quad (13)$$

Denote

$$\rho_1 = \tilde{\boldsymbol{y}}_j^T \cdot \boldsymbol{x}_k, \quad (14)$$
$$\rho_2 = \tilde{\boldsymbol{y}}_j^T \cdot \boldsymbol{x}_{p+k}. \quad (15)$$

First, we examine a simple case, the solution, $(\beta_k^{(1)}, \beta_k^{(2)})$, satisfies

$$\begin{cases} \beta_k^{(1)} > 0, \\ \beta_k^{(2)} > 0, \\ \beta_k^{(1)} < \beta_k^{(2)}. \end{cases} \quad (16)$$

Take derivative of objective function (11), and we have

$$\frac{\partial \tilde{f}}{\partial \beta_k^{(1)}} = \beta_k^{(1)} - \rho_1 + \lambda_1 \text{sgn}(\beta_k^{(1)}) + \lambda_2 \text{sgn}(\beta_k^{(1)} - \beta_k^{(2)}), \quad (17)$$

$$\frac{\partial \tilde{f}}{\partial \beta_k^{(2)}} = \beta_k^{(2)} - \rho_2 + \lambda_1 \text{sgn}(\beta_k^{(2)}) - \lambda_2 \text{sgn}(\beta_k^{(1)} - \beta_k^{(2)}), \quad (18)$$

where $\text{sgn}(\cdot)$ is the sign function.

When $\rho_1 > \lambda_1 - \lambda_2$ and $\rho_2 > \rho_1 + 2\lambda_2$, we have

$$\begin{cases} \beta_k^{(1)} = \rho_1 - \lambda_1 + \lambda_2, \\ \beta_k^{(2)} = \rho_2 - \lambda_1 - \lambda_2. \end{cases} \quad (19)$$

Similarly, we derive all closed-form solutions to problem (10), depending on the values of $\rho_1, \rho_2$ with respect to $\lambda_1, \lambda_2$. The plane $(\rho_1, \rho_2)$ is divided into 13 regions, as shown in Figure 1.

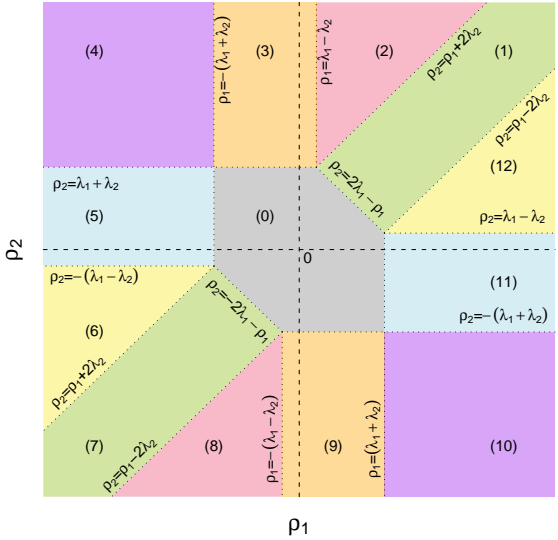

Figure 1: Solution regions of the sub-problem.

Depending on the location of $(\rho_1, \rho_2)$ in the plane, the solutions to problem (10) are as follows.

If $(\rho_1, \rho_2)$ is in region (0), then
$$\beta_k^{(1)} = \beta_k^{(2)} = 0. \tag{20}$$

If $(\rho_1, \rho_2)$ is in region (1), then
$$\beta_k^{(1)} = \beta_k^{(2)} = \frac{1}{2}(\rho_1 + \rho_2) - \lambda_1 \tag{21}$$

If $(\rho_1, \rho_2)$ is in region (2), then
$$\begin{cases} \beta_k^{(1)} = \rho_1 - \lambda_1 + \lambda_2, \\ \beta_k^{(2)} = \rho_2 - \lambda_1 - \lambda_2. \end{cases} \tag{22}$$

If $(\rho_1, \rho_2)$ is in region (3), then
$$\begin{cases} \beta_k^{(1)} = 0, \\ \beta_k^{(2)} = \rho_2 - \lambda_1 - \lambda_2. \end{cases} \tag{23}$$

If $(\rho_1, \rho_2)$ is in region (4), then
$$\begin{cases} \beta_k^{(1)} = \rho_1 + \lambda_1 + \lambda_2, \\ \beta_k^{(2)} = \rho_2 - \lambda_1 - \lambda_2. \end{cases} \tag{24}$$

If $(\rho_1, \rho_2)$ is in region (5), then
$$\begin{cases} \beta_k^{(1)} = \rho_1 + \lambda_1 + \lambda_2, \\ \beta_k^{(2)} = 0. \end{cases} \tag{25}$$

If $(\rho_1, \rho_2)$ is in region (6), then
$$\begin{cases} \beta_k^{(1)} = \rho_1 + \lambda_1 + \lambda_2, \\ \beta_k^{(2)} = \rho_2 + \lambda_1 - \lambda_2. \end{cases} \tag{26}$$

If $(\rho_1, \rho_2)$ in region (7), then
$$\beta_k^{(1)} = \beta_k^{(2)} = \frac{1}{2}(\rho_1 + \rho_2) + \lambda_1. \tag{27}$$

If $(\rho_1, \rho_2)$ is in region (8), then
$$\begin{cases} \beta_k^{(1)} = \rho_1 + \lambda_1 - \lambda_2, \\ \beta_k^{(2)} = \rho_2 + \lambda_1 + \lambda_2. \end{cases} \tag{28}$$

If $(\rho_1, \rho_2)$ is in region (9), then
$$\begin{cases} \beta_k^{(1)} = 0, \\ \beta_k^{(2)} = \rho_2 + \lambda_1 + \lambda_2. \end{cases} \tag{29}$$

If $(\rho_1, \rho_2)$ is in region (10), then
$$\begin{cases} \beta_k^{(1)} = \rho_1 - \lambda_1 - \lambda_2 \\ \beta_k^{(2)} = \rho_2 + \lambda_1 + \lambda_2. \end{cases} \tag{30}$$

If $(\rho_1, \rho_2)$ is in region (11), then
$$\begin{cases} \beta_k^{(1)} = \rho_1 - \lambda_1 - \lambda_2 \\ \beta_k^{(2)} = 0. \end{cases} \tag{31}$$

If $(\rho_1, \rho_2)$ is in region (12), then
$$\begin{cases} \beta_k^{(1)} = \rho_1 - \lambda_1 - \lambda_2 \\ \beta_k^{(2)} = \rho_2 - \lambda_1 + \lambda_2. \end{cases} \tag{32}$$

### 4.3 Convergence analysis

Finally, we summarize the optimization procedure to solve problem (7) in Algorithm 1.

---
**Algorithm 1** Block coordinate descent algorithm to solve problem (7)
---
**Initialization:** $\boldsymbol{\beta}^0 = [0, 0, ..., 0]$, $r = 0$
**while** $\boldsymbol{\beta}^r$ is not converged **do**
    $k \leftarrow (r \mod p) + 1$
    **if** $k \neq j$ **then**
        Let $\beta_l^{(1),r+1} = \beta_l^{(1),r}$, $\beta_l^{(2),r+1} = \beta_l^{(2),r}$, $l \neq k$
        Calculate $\tilde{\boldsymbol{y}}_j$ according to (12).
        Calculate $\rho_1$ and $\rho_2$ using (14) and (15).
        Update $\beta_k^{(1),r+1}$ and $\beta_k^{(2),r+1}$, according to (20)-(32).
    **end if**
    $r \leftarrow r + 1$
**end while**

---

The convergence of Algorithm 1 is stated in the following theorem.

**Theorem 1.** *The solution sequence generated by Algorithm 1 is bounded and every cluster point is a solution of problem* (7).

*Proof.* We have shown in Property 1 and Property 2 that the objective function (6) is continuous and convex, and the non-differential part of the objective function is block-wise separable. By applying Theorem 4.1 proposed by Tseng (2001), we have that the solution sequence generated by Algorithm 1 is bounded and every cluster point is a solution of problem (7). □

### 4.4 Determining parameters $\lambda_1$ and $\lambda_2$

As we discussed previously, the first $\ell_1$-regularization term, $\lambda_1 \|\boldsymbol{\beta}\|_1$, leads to the identification of sparse graph structures, and the second $\ell_1$-regularization term, $\lambda_2 \|\boldsymbol{\beta}^{(1)} - \boldsymbol{\beta}^{(2)}\|_1$, suppresses the inconsistencies of the network structures and parameters between two conditions, due to the noise in the data and limited samples.

First, we consider the case $\lambda_2 = 0$. In this case, the problem (7) is equivalent to applying lasso to the data under two conditions separately. The $\lambda_1$ controls the sparsity of the learned graph, and Algorithm 1 is reduced to a coordinate descent algorithm, in which each sub-problem is lasso with two orthogonal predictors. The value of $\lambda_1$ can be determined easily via cross-validation. In our experiments, we used 10-fold cross-validation, following steps specified in (Hastie *et al.*, 2008).

Then we consider the second parameter $\lambda_2$. The parameter $\lambda_2$ controls the sparsity of structural and parametric changes between two conditions. From regions (1) and (7) of Figure 1, we can see that if $|\rho_1 - \rho_2| \leq 2\lambda_2$, then $\beta_k^{(1)}$ and $\beta_k^{(2)}$ will be set equal (Equations (21) and (27)) as the solution of the sub-problem (10). Therefore, the remaining question is when $|\rho_1 - \rho_2|$ is large enough to be considered significant, at a given significance level $\alpha$. We present here a heuristic approach to determine $\lambda_2$.

Applying Fisher transform to both $\rho_1$ and $\rho_2$, we have

$$z_1 = \frac{1}{2} \ln \frac{1+\rho_1}{1-\rho_1}, \quad z_2 = \frac{1}{2} \ln \frac{1+\rho_2}{1-\rho_2}. \quad (33)$$

Since data matrices $\mathbf{X}_1$ and $\mathbf{X}_2$ are drawn from Gaussian distributions, we know $z_1$ and $z_2$ are approximately normally distributed with standard deviation $\frac{1}{\sqrt{N-3}}$ and means $\frac{1}{2} \ln \frac{1+\rho_1}{1-\rho_1}$ and $\frac{1}{2} \ln \frac{1+\rho_2}{1-\rho_2}$, respectively.

Further, under the null hypothesis that $\rho_1 = \rho_2$ (and therefore $z_1 = z_2$), define

$$z = z_1 - z_2, \quad (34)$$

which approximately follows normal distribution with zero mean and standard deviation $\frac{1}{\sqrt{(N-3)/2}}$.

At a given significance level $\alpha$ (*e.g.*, $\alpha = 0.01$ is used in Section 5), if $|z| = |z_1 - z_2| \geq s$, it will be considered significant, where $s = \Phi^{-1}(1-\alpha/2)/\sqrt{(N-3)/2}$. Through simple derivation, we have

$$|z| = |z_1 - z_2| \geq s$$
$$\Rightarrow |\rho_1 - \rho_2| \geq \frac{e^{2s} - 1}{e^{2s} + 1}(1 - \rho_1\rho_2) = 2\lambda_2 \quad (35)$$

To further simplify (35) with some approximation, we estimate overall $\rho_1\rho_2$ by $\overline{\rho_1\rho_2} = 2\sum_{j<l} \boldsymbol{y}_j^T \boldsymbol{x}_l \cdot \boldsymbol{y}_j^T \boldsymbol{x}_{p+l}/p(p-1)$. Substituting $\overline{\rho_1\rho_2}$ in (35), we have

$$\lambda_2 = \frac{e^{2s} - 1}{2e^{2s} + 2}(1 - \overline{\rho_1\rho_2}). \quad (36)$$

## 5 Experiments

### 5.1 A synthetic experiment

We first use a synthetic example to illustrate the principle and test the proposed method. Assume there are six nodes in the Gaussian graphical model, $A$, $B$, $C$, $D$, $E$, $F$. Under condition 1, their relationships are represented by Figure 2a. Under condition 2, their relationships are altered, as shown in Figure 2b. We generated 200 samples from the joint Gaussian distribution according to the Gaussian graphical model with the structure specified by Figure 2a, and 200 samples from the joint Gaussian distribution according to Gaussian graphical model with the structure specified by Figure 2b.

The penalty parameters are set to $\lambda_1 = 0.22$ and $\lambda_2 = 0.062$, calculated according to Section 4.4. Figure 3a is the composite network under two conditions inferred by the proposed algorithm, where the black lines are the edges that exist under both conditions, the red lines are the edges that exist only under condition 1 and the green lines are the edges that exist only under condition 2. Since we are more interested in the changed part of the graph, we extracted the edges and nodes involved in the changes to highlight these structural changes. We term it differential sub-network, as shown in Figure 3b. We can see the proposed algorithm accurately captured the structural changes of the graphical model between two conditions.

### 5.2 Experiment on modeling gene regulatory networks under two conditions

Inference of the structures of gene regulatory networks from expression data is a fundamental problem in com-putational biology. Our goal here is to infer and extract the structural changes of a gene regulatory network between two conditions using gene expression data. SynTReN is a network generator that creates synthetic transcriptional regulatory networks and produces simulated gene expression data that approximate experimental data, used as benchmarks for the validation of bioinformatics algorithms (Van den Bulcke *et al.*, 2006).

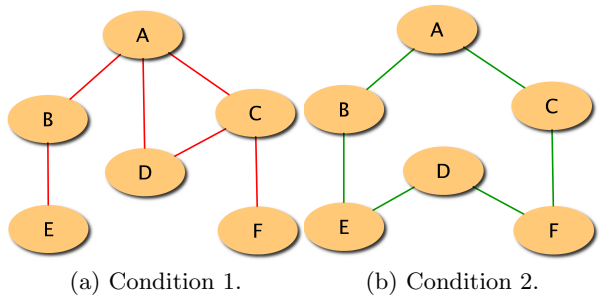

(a) Condition 1.   (b) Condition 2.

Figure 2: The structures of the Gaussian graphical model under two conditions.

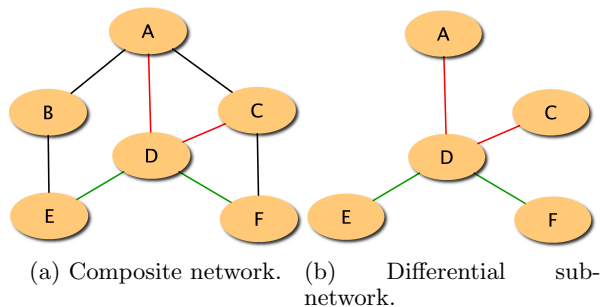

(a) Composite network.   (b) Differential sub-network.

Figure 3: The network structure learned by the proposed method. The black lines are the edges that exist under both conditions. The red lines are the edges that exist only under condition 1. The green lines are the edges that exist only under condition 2.

To test the applicability of the proposed framework in gene regulatory network modeling, we used the software SynTReN to generate one simulation dataset of 50 samples of a sub-network drawn from an existing signaling network in *Saccharomyces cerevisiae*. Then we changed part of network and used SynTReN to generate another dataset of 50 samples according to this modified network. The networks under two conditions is shown in Figure 4a. The network contains 20 nodes that represent 20 genes. The black lines indicate the regulatory relationships that exist under both conditions. The red and green lines are the regulatory relationships that exist only under condition 1 and condition 2, respectively. The sub-network comprised of nodes MBP1_SWI6, CLB5, CLB6, PHO2, FLO1, FLO10 and TRP4 and the green and red lines is the focus of our study that our algorithm tries to identify from expression data.

Figure 4b shows the differential sub-network between the two conditions extracted by the proposed algorithm. The penalty parameters are set to $\lambda_1 = 0.28$ and $\lambda_2 = 0.123$, calculated according to Section 4.4. Compared with the known network topology shown in Figure 4a, the proposed algorithm correctly identified all the nodes with structural changes and 7 of 10 differential edges. The edge between CDC10 and ACE2 was falsely detected. This indicates that our algorithm can successfully detect these interesting genes using their network structure information, even though the means of their expressions did not change substantially between the two conditions. Therefore, this method is able to identify biomarkers that cannot be detected by traditional gene ranking methods, providing a complimentary approach for biomarker identification problem.

## 6 Conclusions

In this paper, we reported an effective learning strategy to extract structural changes in Gaussian graphical models in controlled experiments. We presented a convex optimization framework using $\ell_1$-regularization to formulate this problem, and introduced an efficient block coordinate descent algorithm to solve it. We demonstrated the effectiveness of the approach on a numerical simulation experiment, and then we applied the algorithm to detecting gene regulatory network structural changes under two conditions and obtained very promising results.

#### Acknowledgements

This work was supported in part by the National Institutes of Health under Grants NS029525 and CA149147.

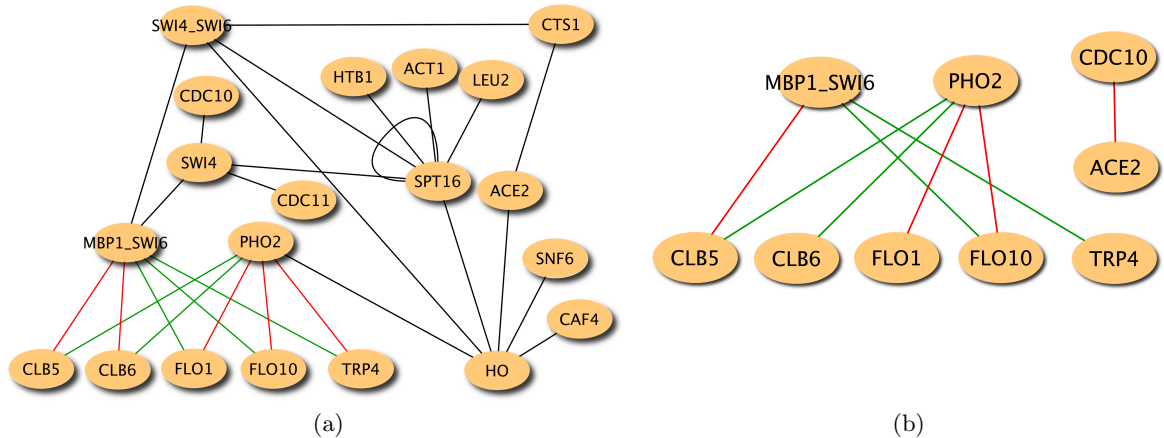

Figure 4: (a) The gene regulatory network under two conditions. Nodes in the network represent genes. Lines in the network indicate regulatory relationships between genes. The black lines are the regulatory relationships that exist under both conditions. The red and green lines represent the regulatory relationships that exist only under condition 1 and under condition 2, respectively. (b) The sub-network extracted by the proposed algorithm.